\documentclass{article}

\PassOptionsToPackage{numbers, compress}{natbib}

\usepackage[final]{neurips_2019}

\usepackage[utf8]{inputenc} 
\usepackage[T1]{fontenc}    
\usepackage{hyperref}       
\usepackage{url}            
\usepackage{booktabs}       
\usepackage{amsfonts}       
\usepackage{nicefrac}       
\usepackage{microtype}      
\usepackage{epsfig}
\usepackage{graphicx}

\newcommand*\samethanks[1][\value{footnote}]{\footnotemark[#1]}

\title{Image Based Identification of Ghanaian Timbers Using the XyloTron: Opportunities, Risks and Challenges}

\author{Prabu Ravindran, \textsuperscript{1, 2}
		\And
	    Emmanuel Ebanyenle\thanks{Part of the work done at the USDA Forest Products Laboratory, USA as Visiting Scientists}, \textsuperscript{3}
	    \And
		Alberta Asi Ebeheakey\samethanks, \textsuperscript{4}
		\And
		Kofi Bonsu Abban\samethanks, \textsuperscript{4}
		\And
		Ophilious Lambog\samethanks, \textsuperscript{4}
		\And
		Richard Soares, \textsuperscript{1, 2}
		\And
		Adriana Costa, \textsuperscript{1}
		\And
		Alex C. Wiedenhoeft \textsuperscript{1, 2, 5, 6}\\
	\textsuperscript{1}{Center for Wood Anatomy Research, USDA Forest Products Laboratory, USA}\\
	\textsuperscript{2}{Department of Botany, University of Wisconsin, Madison, USA}\\
	\textsuperscript{3}{Wood Anatomy Laboratory, CSIR - Forestry Research Institute of Ghana}\\
	\textsuperscript{4}{Timber Industry Development Division, Forestry Commission, Ghana}\\
	\textsuperscript{5}{Department of Forestry and Natural Resources, Purdue University, USA}\\
	\textsuperscript{6}{Ciências Biolôgicas, Universidade Estadual Paulista – Botucatu, Brasil}}

\begin{document}

\maketitle

\begin{abstract}
 	Computer vision systems for wood identification have the potential to empower both producer and consumer countries to combat illegal logging if they can be deployed effectively in the field. In this work, carried out as part of an active international partnership with the support of UNIDO, we constructed and curated a field-relevant image data set to train a classifier for wood identification of $15$ commercial Ghanaian woods using the XyloTron system.  We tested model performance in the laboratory, and then collected real-world field performance data across multiple sites using multiple XyloTron devices. We present efficacies of the trained model in the laboratory and in the field, discuss practical implications and challenges of deploying machine learning wood identification models, and conclude that field testing is a necessary step - and should be considered the gold-standard - for validating computer vision wood identification systems.  
\end{abstract}

\section{Introduction}
Illegal logging contributes to deforestation and environmental degradation, supports organized crime networks, and its negative financial impact is valued between US$\$50-150$ billion~\cite{Unep2012}. To combat this, there is growing global interest in enacting and enforcing laws (e.g. CITES, Lacey Act) intended to ensure that wood and wood-derived products are legally sourced. Compliance with and enforcement of international and local laws for legal wood products depend in part on the availability of technical or forensic expertise to validate claims of legality~\cite{Wiedenhoeft2019}. Such expertise in turn hinges on the design, validation, and deployment of robust scientific wood forensic pipelines to identify timber and combat fraud throughout the supply chain~\cite{LoweSasaki2016}.  

In a typical scenario, adversarial operators falsify paperwork claiming that the wood in a consignment is of lower value when in reality the consignment contains higher-value/endangered, (sometimes) superficially similar species. Verifying a consignment claim amounts to making a correct identification of the timber based solely on its inherent characteristics in the context of the claimed species. In field screening, the inspector must identify the wood in uncontrolled environmental conditions (e.g.~at the point of harvest, in a lumber mill, at the harbor) in a matter of seconds to establish probable cause for seizure, detention, and further forensic analyses, or release the consignment into trade as compliant. In most jurisdictions, specimens from a detained consignment will be subjected to further forensic analysis in a laboratory using genetic/microscopy/spectral techniques to enable a legally valid identification~\cite{Dormontt2015}.    

The \emph{de facto} state of the art in field screening of timber in most of the world is human-based, with effective inspectors requiring significant training and regular practice in the use of traditional wood anatomical identification methods.  Such trained humans will typically restrict their inspection of a wood specimen to the knife-cut transverse surface (the end grain) in order to view the size, shape, abundance and relationships of the constituent cells using a hand lens. Due to the comparative dearth of such human expertise in most countries, field screening of timber, \emph{if done at all}, more often relies on subjective features such as color or odor of the timber with no reference to anatomical features of the wood. The lack of sufficient human expertise compared to the demand for timber screening is a major bottleneck in ensuring legal timber trade, and has established the clear need for reliable field-deployable wood identification technologies~\cite{Wiedenhoeft2019}.   

\begin{figure*}[h]
	\centering
	\includegraphics[width=0.85\textwidth,height=0.15\textheight]{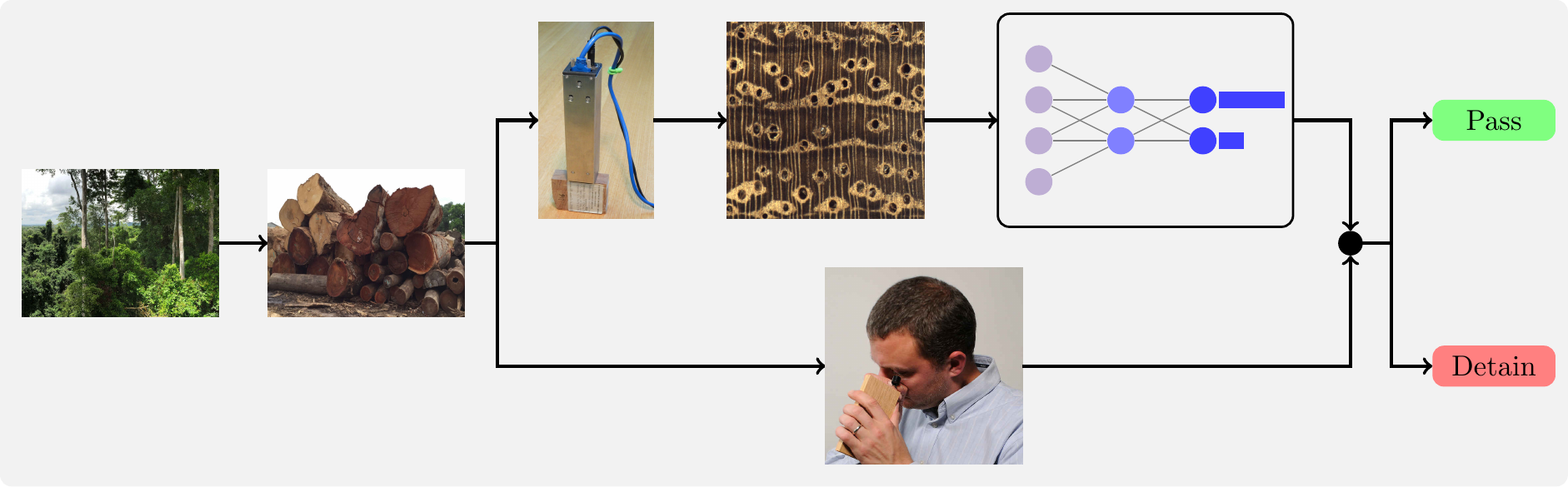}
	\caption{Context and workflow for field screening of timbers. The top branch represents the computer vision/machine learning approach.  The bottom branch represents the traditional human based approach. Both approaches converge to inform a decision to detain or pass a shipment.}
	\label{fig:feature-graphic}
\end{figure*}

Computer vision and machine learning are attractive technologies for the development of quick, reliable and field-deployable tools for field screening of wood~\cite{HermansonWiedenhoeft2011}. Image-based identification of a wood specimen using field-collected images of the transverse surface of wood is similar to the well-studied problem of texture classification~\cite{TexClassSurvey2018, CimpoiDeepTexture2015}. An early work using machine learning for wood identification used handcrafted biometric measurements and descriptors with a multi-layer perceptron to distinguish between two species~\cite{EstebanANN2009}. In~\cite{Khalid2008} a wood identification system was developed using gray level co-occurence matrices~\cite{Haralick1979} and multi-layer perceptrons. Local binary patterns were used to identify African timber species using microscopic images in~\cite{AfricaMicroID2017}. Convolutional neural networks~\cite{LeCun1989} were employed to automatically learn features for macroscopic wood species identification in~\cite{Filho2014, CostaRicaCutting2018} and were designed for laboratory settings. The work of~\cite{Ravindran2018} uses transfer learning~\cite{PanTransferLearnSurvey2010, ZamirTaskonomy2018}, with a pretrained VGG network~\cite{Simonyan2014} to identify neotropical woods from the Meliaceae, the botanical family that includes the genuine mahoganies. 

A critical aspect not addressed in prior literature on computer vision wood identification is "ground-truth" testing of real-world performance in the context of field deployment by the personnel responsible for adoption and application of the technology.  Based on the variability of wood itself and the need to prepare wood for imaging in the field, it would be optimistic to the point of naivete to assert or assume that test data set performance  would direcly translate to the real world.  In this regard, computer vision wood identification suffers from many of the same constraints as trained human inspectors. In addition to practical concerns for preparing specimens, the system's user interface and the mechanisms for reporting classification outcomes to the operator is expected to influence the adoption and utility of the technology. Developing a system and user interface that delivers the relevant and digestible granularity of information is central to the power of the technology.  A tool must empower its user to perform their work more effectively, or to enable the user to take on new tasks not previously possible. To this end, it is critical for computer vision researchers to work with end users to ensure that all necessary - and no extraneous - detail is conveyed to the user in a format that is empowering rather than confusing or opaque.

In this paper we use transfer learning to train a ResNet~\cite{HeResNet2015} based classifier for image-based, macroscopic identification of a subset of commercially important Ghanaian timbers for use in conjunction within the XyloTron system~\cite{XyloScope}. Our data collection and model development was done as part of an active UNIDO-funded international partnership to improve timber tracking and timber forensics in the Ghanaian timber market using xylarium wood specimens from the US Forest Products Laboratory and the Forestry Research Institute of Ghana. The pilot study described here was tested both in the laboratory and in the field to yield valuable insights into the challenges to be encountered when scaling the number of taxa to be identified. To the best of our knowledge this is the first report of results of field testing a computer vision/machine learning model for wood identification.

\section{Dataset}

\subsection{Sample preparation and imaging} 
The transverse surfaces of $413$ xylarium specimens of $38$ species in $15$ genera of commercial interest in Ghana were prepared in order to make the anatomical features of the wood easier to visualize. The list of taxa used in the created dataset are listed in Table~\ref{tab:comp_split}. 

\begin{figure*}
	\centering
	\includegraphics[width=0.95\textwidth,height=0.35\textheight]{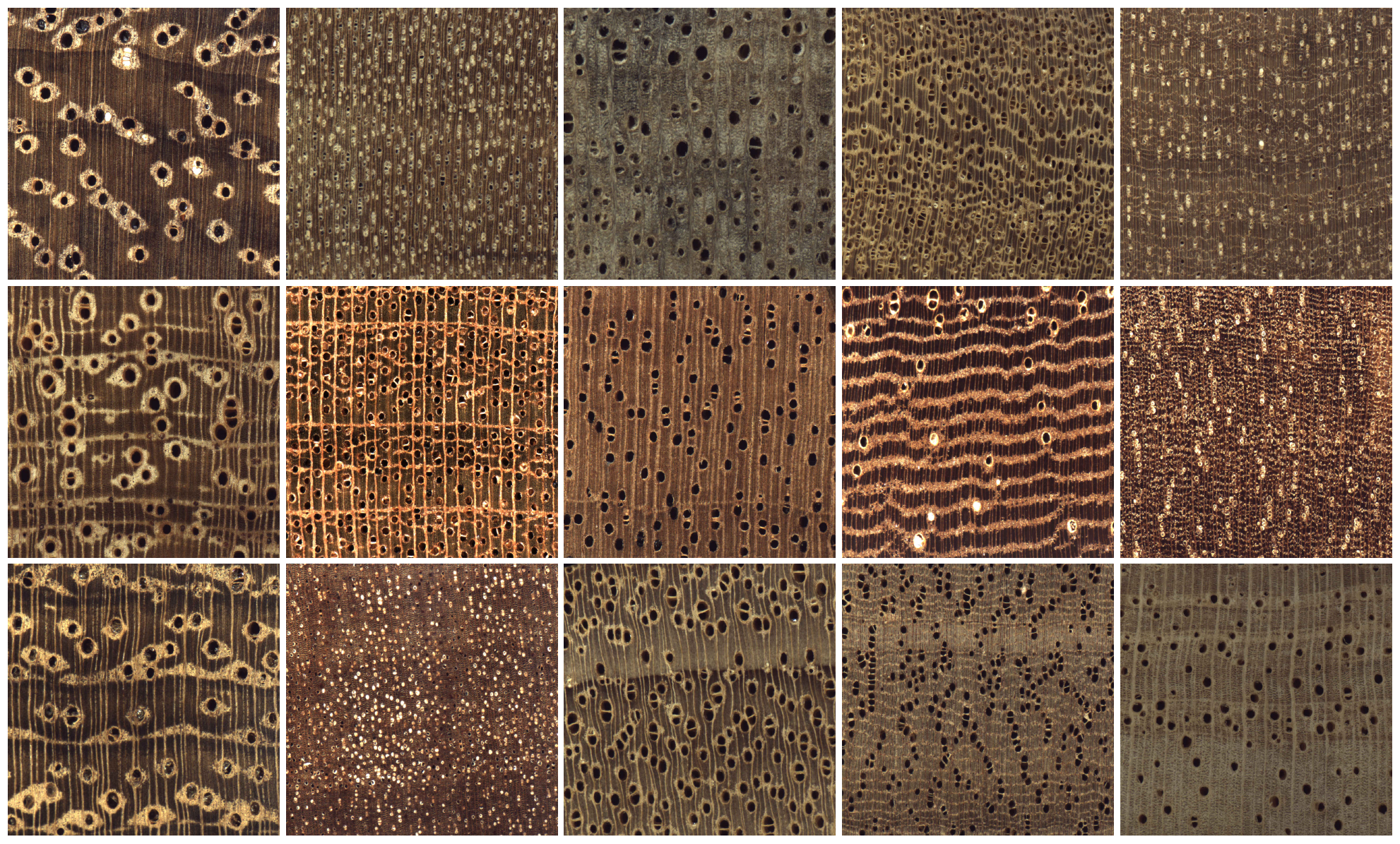}
	\caption{Archetypal example images for the classes. The classes in table~\ref{tab:comp_split} are displayed in raster order. \emph{A single image for each class cannot represent the natural intra-class wood anatomical variability.}}
	\label{fig:mosaic}
\end{figure*}

Macroscopic images of the prepared surfaces were obtained using the XyloTron, a DIY, open-source macroscopic imaging system~\cite{XyloScope}. Multiple \emph{non-overlapping} images were captured from each specimen with the rays of the wood aligned vertically. Vertical ray alignment was for consistency with existing scientific image collections and for ease of human-mediated interventions. Exemplar images for the classes considered in this paper are shown in Figure~\ref{fig:mosaic}.

\subsection{Data curation}
The collected images were curated by a wood forensic expert and images showing atypical wood anatomy or misidentified wood specimens were removed from the dataset. This resulted in a total of $2187$ images. The class labels were assigned so that the species in Table~\ref{tab:comp_split} would be identified/classified at the genus level. This genus-level granularity for the labels is consistent with capabilities of traditional wood anatomy based timber identification and the commercial demands of the timber market in Ghana. Additionally, in order to increase the number of exemplars per class, when needed we included images of macroscopically similar species of the same genus from outside Ghana. The species compositions for our $15$ classes are listed in Table~\ref{tab:comp_split}.

\begin{table}[]
	\begin{centering}
		\begin{tabular}{|l|c|c|c|}
			\hline
			Class & 
			Species Composition &
			\begin{tabular}{@{}c@{}}
				\emph{Specimen counts} \\ 
				\emph{(train, valid, test)}
			\end{tabular} &
			\begin{tabular}{@{}c@{}}
				\emph{Image counts} \\ 
				\emph{(train, valid, test)}
			\end{tabular}\\ 
			\hline
			\hline
			Albizia	& 
			\begin{tabular}{@{}c@{}}
				\emph{A. adianthifolia}\\
				\emph{A. ferruginea}\\
				\emph{A. zygia} 
			\end{tabular} & 
			(21, 5, 5) & 
			(141, 30, 30) \\
			\hline
			Canarium & 
			\emph{C. schweinfurthii} & 
			(7, 2, 2) & 
			(61, 13, 13)  \\
			\hline
			Ceiba & 
			\emph{C. pentandra} & 
			(25, 6, 6) & 
			(140, 30, 30)\\
			\hline
			Celtis &   
			\begin{tabular}{@{}c@{}} 
				\emph{C. adolfi-friderici}\\ 
				\emph{C. mildbraedii}\\
				\emph{C. zenkeri}	
			\end{tabular} & 
			(4, 3, 3) & 
			(90, 16, 21)\\
			\hline 
			Chrysophyllum &  
			\begin{tabular}{@{}c@{}}
				\emph{C. albidum}\\
				\emph{C. brieyi}\\
				\emph{C. fulvum}\\
				\emph{C. lacourtianum}\\ 
				\emph{C. perpulchrum}\\
				\emph{C. viridifolium}
			\end{tabular} & 
			(5, 1, 1) & 
			(77, 16, 16)\\
			\hline 
			Daniellia & 
			\begin{tabular}{@{}c@{}}
				\emph{D. ogea}\\
				\emph{D. oliveri} 
			\end{tabular} & 
			(3, 2, 2) & 
			(37, 12, 12)\\
			\hline
			Entandrophragma & \begin{tabular}{@{}c@{}}
				\emph{E. angolense}\\ 
				\emph{E. candollei}\\
				\emph{E. cylindricum}\\
				\emph{E. utile} 
			\end{tabular} & 
			(60, 14, 14) & 
			(141, 30, 30)\\
			\hline
			Khaya &  
			\begin{tabular}{@{}c@{}}
				\emph{K. anthotheca}\\
				\emph{K. ivorensis}\\ 
				\emph{K. senegalensis} 
			\end{tabular} & 
			(43, 9, 9) & 
			(142, 30, 30)\\
			\hline
			Lophira	& 
			\emph{Lophira alata} & 
			(5, 1, 1) & 
			(60, 11, 12)\\
			\hline
			Manilkara &  
			\begin{tabular}{@{}c@{}}
				\emph{M. bidentata}\\
				\emph{M. elata}\\
				\emph{M. huberi}\\
				\emph{M. obovata}\\
				\emph{M. solimoesensis}\\ 
				\emph{M. zapotilla} 
			\end{tabular} & 
			(44, 10, 10) & 
			(142, 29, 29)\\
			\hline
			Milicia &  
			\begin{tabular}{@{}c@{}}
				\emph{M. excelsa}\\
				\emph{M. regia} 
			\end{tabular} & 
			(18, 3, 3) & 
			(141, 29, 29)\\
			\hline 
			Nesogordonia & 
			\emph{N. papaverifera} & 
			(7, 1, 1) & 
			(67, 15, 14)\\
			\hline  
			Terminalia &   
			\begin{tabular}{@{}c@{}}
				\emph{T. ivorensis}\\ 
				\emph{T. superba} 
			\end{tabular} & 
			(24, 5, 5) & 
			(141, 30, 30)
			\\
			\hline
			Tieghemella &  
			\begin{tabular}{@{}c@{}}
				\emph{T. africana}\\
				\emph{T. heckelii} 
			\end{tabular} & 
			(6, 2, 2) & 
			(65, 13, 14)\\
			\hline
			Triplochiton & 
			\emph{T. scleroxylon} & 
			(9, 2, 2) & 
			(90, 19, 19)\\
			\hline
			\hline
		\end{tabular}
	\end{centering}
	\caption{
		Botanical species composition of the $15$ genus-level classes.}
	\label{tab:comp_split}
\end{table}

\section{Model training and deployment}
\subsection{Model architecture}
\begin{figure*}[t]
	\centering
	\includegraphics[width=\textwidth]{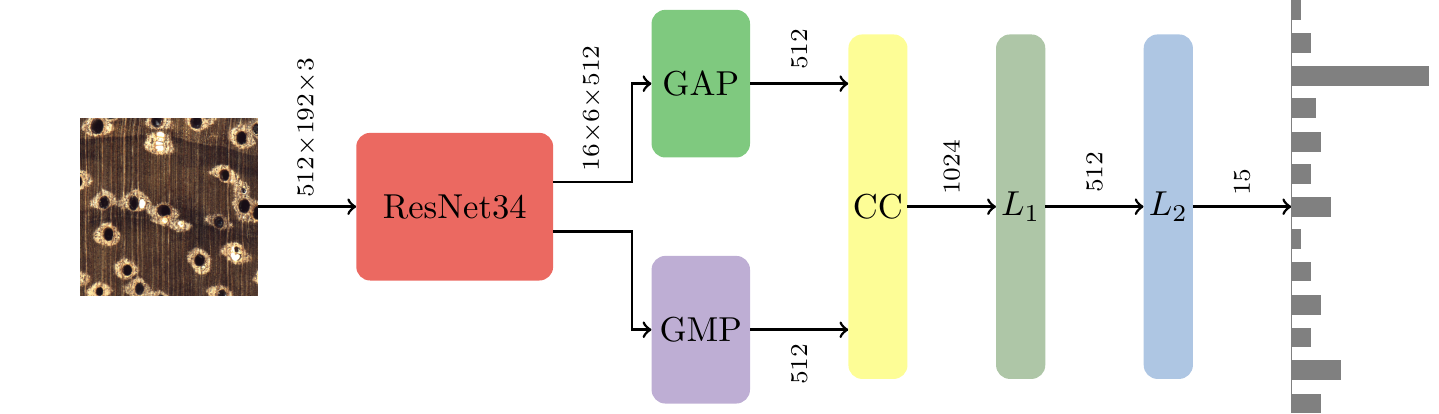}
	\caption{The model backbone comprises all the residual blocks of an ImageNet-pretrained ResNet$34$. We use a custom head as shown. GAP: Global Average Pooling, GMP: Global Max Pooling, CC: Concatenation, $L_1$: BatchNorm-Dropout(0.25)-Linear-ReLU and $L_2$: BatchNorm-Dropout(0.5)-Linear-SoftMax.}
	\label{fig:arch}
\end{figure*}
Our model comprises all the layers up to and including the final residual block from a ResNet34~\cite{HeResNet2015} network pre-trained on ImageNet~\cite{Russakovsky2015}. To this ResNet backbone we added concatenated global max and average pooling layers and two Batchnorm~\cite{IoffeSzegedyBN2015}-Dropout~\cite{HintonDropout2012}-Linear (BDL) blocks. ReLU~\cite{Nair2010} activation and a dropout probability of $0.25$ was used in the first BDL block. In the second BDL block a dropout probability of $0.5$ and a softmax activation with $15$ outputs was used. The CNN architecture is shown in Figure~\ref{fig:arch}.

\subsection{Data splits}
We divided the dataset of $2187$ images into a (approximately) $70\%/15\%/15\%$ training/validation/testing split with every specimen contributing images to \emph{exactly} one of the splits. The number of specimens per class and images per specimen were not constant, but varied according to the availability of correctly identified specimens and the cross-sectional (transverse) area of those specimens. We employed stratified sampling at the specimen level to generate the splits, the details of which are shown in Table~\ref{tab:comp_split}.    

\subsection{Training methodology}
The backbone ResNet layers were initialized with ImageNet~\cite{Russakovsky2015} pretrained weights and He normal initialization~\cite{HeInit2015} was used for the the custom top level layers. We use a typical two-stage transfer learning process to train our model. In the first stage, the ResNet backbone was used as a feature extractor~\cite{Razavian2014} (i.e. layer weights frozen) and the custom top level was trained for $6$ epochs using the Adam optimizer. In the second stage of training we finetuned the weights of the entire network for $8$ epochs using Adam optimizer~\cite{KingmaAdamOpt2014}. The number of epochs to run was estimated using the validation set.

During both the training stages we simultaneously annealed the learning rate and momentum as described in~\cite{SmithDisciplined2018, howard2018fastai}. For the first half of the annealing process, the learning rate was increased from $\alpha_{\mbox{\small min}}$ to $\alpha_{\mbox{\small max}}$ while simultaneously the momentum was decreased from $\beta_{\mbox{\small max}}$ to $\beta_{\mbox{\small min}}$. In the second half the learning rate was decreased while the momentum was increased. Cosine annealing was used throughout. For the first and second stages of training the value of $\alpha_{\mbox{\small max}}$ was set to $2e^{-2}$ and ${1e^{-5}}$ respectively and was estimated using the method in~\cite{SmithCyclical2017}. The value of  $\alpha_{\mbox{\small min}}$ was set to  ${\alpha_{\mbox{\small max}}}/10$. The momentum parameters $\beta_{\mbox{\small min}}$ and $\beta_{\mbox{\small max}}$ were set to $0.85$ and $0.95$ respectively.  

We extracted patches of size $2048 \times 768$ pixels and resized them to $512 \times 192$ pixels. The reason we extracted non-square patches was to ensure that the diagnostic wood anatomical growth transitions were maximally probable to be captured in every patch. Our data augmentation strategy included horizontal and vertical reflections, random rotations in the range $[-5, 5]$ degrees and cutout~\cite{Cutout2017}. The architecture definition and training was implemented using PyTorch and scientific Python~\cite{Scipy2014} on a NVIDIA Titan X GPU using a batch size of $16$.     

\subsection{Hardware and software for field deployment}
Four XyloTrons, each consisting of a XyloScope~\cite{XyloScope} for imaging and an off-the-shelf laptop for running inference, were deployed in Ghana for model evaluation.  The user interface displayed the image being classified along with the top-$3$ prediction confidences and archetypal images for these predictions from the reference image set (Figure~\ref{fig:mosaic}). Displaying an archetypal image for the prediction alongside the unknown specimen allows the operator to incorporate an element of human validation into the process, and reinforce the operator's knowledge of the anatomy of the woods in the model.

\section{Results and discussion}
\subsection{Laboratory testing results}
The overall image-level top-$1$ accuracy of our model was $97\%$. In our experiments finetuning did not improve the performance of the model.  The few incorrect predictions made by our model are consistent with kinds of errors that would be made by a trained human inspector using a hand lens to identify the timbers using the same anatomical detail available in our image data set. For example, ${3}\%$ of \emph{Ceiba} test images were classified as \emph{Triplochiton}, which is reasonable based on both wood anatomy (see Figure~\ref{fig:mosaic}) and their shared botanical family, the Malvaceae. Within the family Meliaceae, some images of \emph{Entandrophragma} were classified as \emph{Khaya}, which is also reasonable given the anatomical variability of these two woods. \emph{Tieghemella} and \emph{Chrysophyllum} both belong to the Sapotaceae, a family known for taxonomic lability and difficulty of separating species and even genera when observing a standing tree with bark and leaves. Given the similarities in wood anatomy and underlying botanical variation for these three predictions, the error rates are much lower than one would expect for anyone other than an expert forensic wood anatomist.  We would like to emphasize that the XyloTron uses a single image of the transverse surface of an unknown specimen for prediction, whereas a trained human identifier would incorporate multiple fields of view and additional anatomical information from other surfaces of the unknown specimen.

\subsection{Field testing results}
The model was deployed at three field locations in Ghana, and at the xylarium at the Forestry Research Institute of Ghana. Across these locations a total of $488$ specimens were evaluated using four different XyloTron units by multiple users. In this pilot evaluation of field deployment, the operators were asked to record and report the number of times each species was correctly identified and if incorrect what the prediction of the model was. The overall accuracy of the model in field testing was $72\%$. There were two broad types of misclassification - those images that were classified as other anatomically similar woods, and one class (Canarium) where the predictions are wrong and not clearly explainable by observable wood structure. 

\subsection{Discussion}
The generalizability of state-of-the-art ImageNet classifiers was recently studied~\cite{Recht2018, Recht2019} by testing the predictive performance of pretrained models on a new dataset with carefully chosen images that were similar to the original test dataset. It was shown that models based on the ResNet architecture had a $\sim10\%$ drop in predictive accuracy on the new dataset and the authors attributed the drop in accuracy (consistent across a broad range of models) to ``\emph{the models' inability to generalize to slightly "harder" images than those found in the original test sets}~\cite{Recht2019}". In our field tests the model was evaluated in an unrestricted setting where the wood specimens were from trees of varying maturity (reported by the field testing personnel). Given this uncontrolled setting, it is likely that the distribution of anatomical characteristics in the field tested specimens was different from the laboratory testing dataset (collected in xylaria) and thus were "harder" for the classifier. The generalizability of this approach across the developing world depends largely on access to sufficient reference specimens and commitment from local users and developers of the technology so that the final, deployed version reflects appropriate scientific, practical, and cultural factors to maximize the technology’s potential contribution to natural resource management. A necessary piece of this is the real-world field testing so that any differences in performance between the laboratory and the field are well characterized. 
  
The gap between test data metrics and real-world performance should be viewed in light of the two common practices currently in place in the field: no testing of any kind, or subjective evaluation of unreliable features such as color and odor. Additionally, this disparity between overall test set accuracy and the field-testing accuracy is a sobering but informative result because it demonstrates how test data set accuracy metrics may not translate into field accuracy metrics and exposes the challenges that biological variability and data distribution mismatches between model training/testing and field deployment specimens can pose for machine learning based wood identification systems. We hope that this work raises awareness of the potential inadequacy of machine learning based wood identification systems that have not been "ground-truthed" in the field. 

\section{Challenges, risks and opportunities}
The vibrant interplay of wood anatomy, computer vision/machine learning and law enforcement/compliance provides a slew of research challenges with opportunities to make real-world impacts in wood utilization, forest management and land-use policy, and conservation of biodiversity. 

\subsection{Scaling up}
A key opportunity in deploying an image-based machine learning wood identification tool in parts of the developing world is the ability to limit the number of classes (different woods) that the system must be trained to separate. For example, Ghana imports virtually no exotic timbers, thus a system trained to identify Ghanaian woods provides real-world value. Deploying a field-screening system in Ghana, that actively uses a timber tracking system, has been an ideal case study to showcase the potential of this approach. In net-importing (and typically developed) countries, a shipment of wood could come from anywhere in the world, and thus an effective system would be required to identify several hundreds, instead of tens, of woods. 

\subsection{Reference/baseline image data sets}
Compared to the global scale of illegal logging, there is a paucity of high quality wood image datasets, in part because there are comparatively few xylaria. Xylaria can be valuable sources of curated wood specimens, but the quality, size, breadth, and reliability of these collections are highly variable~\cite{CostaRicaXyl2018}.  Other options for acquiring wood specimens such as targeted field expeditions, active timber harvest sites, lumber mills, etc. may faithfully capture the current data distribution but can be logistically challenging to accomplish at scale. Clearly, a global, open-access database of large numbers of images of all woods would be ideal, but such an effort would remain limited by specimen access, funding, and expertise. It may be more important- and achievable- to establish a baseline wood image dataset to objectively measure domain-specific machine learning advances - \emph{a la} MNIST of woods. 

\subsection{Machine learning models and computer vision hardware}
We anticipate that advances in computer vision and machine learning will lead to continued improvements for timber screening for the the foreseeable future. Increased access to large datasets may provide the potential to develop more robust models that capture both overt and subtle variation in wood anatomy. Scaling up models to incorporate more classes can be a challenge and may require label space engineering using custom ontologies to handle similar wood anatomies and data scarcity challenges. Techniques like few-shot learning~\cite{KochOneShot2015, VinyalsOneShot2016, ZemelFewShot2017} hold promise for wood discrimination models, especially for endangered, rare endemic, or species of emerging commercial interest where access to new specimens maybe limited. Cloud based deployment of timber screening technologies can be implemented to enable real-time expert mediated decisions~\cite{Tay2017}, but the sites and contexts where they can be reliably deployed are biased toward regions with reliable, high-speed network connectivity. Portable, self-contained systems that do not require real-time network access for cloud based processing, like ours, can be deployed in regions without reliable access to these resources. Regardless of the details of hardware and location of computation, objective evaluation of different models and hardware configurations is only meaningful if compared across a common data set, but even this may not capture factors that facilitate real-world adoption and implementation of the technology.

\subsection{Context-aware implementation and evaluation}
Because compliance with or enforcement of legal logging laws is inherently a human-mediated endeavor that varies by jurisdiction, machine learning technologies can solve only as much of the problem as users are willing and able to adopt and apply the technologies. Developing user-interfaces that mediate access to the model results at the correct and useful level of detail for the application will be a central part of long-term adoption. In some cases, a user interface that provides simple yes/no results may be desirable, whereas in others the interface might provide guidance on an optimal testing scheme according to the uncertainties in predictions.  Which user interface is most useful will depend on the details of the problem and the norms of local jurisprudence, and these factors should be taken into account when deploying and interpreting the efficacy of machine learning models. 

\subsection{The necessity of field testing}
The central conclusion of our work is that field testing is necessary and should be the gold-standard by which computer vision wood identification systems must be evaluated. Any claim about efficacy not backed by field test data at best represents an optimistic projection, and at worst grossly overpromises and underdelivers, with the cost of a gap in performance falling disproportionately on the implementing country. To realize the potential of computer vision (or any other technique) to combat illegal logging, it is essential to direct limited scientific and deployment resources to those field-tested approaches that will have the greatest real world impact.

{\small
	\bibliographystyle{unsrt}
	\bibliography{egbib}

\begin{thebibliography}{10}

\bibitem{Unep2012}
{\em Nellemann, C. and INTERPOL Environmental Crime Programme: Green Carbon,
  Black Trade: Illegal Logging, Tax Fraud, and Laundering in the World's
  Tropical Forests: A Rapid Response Assessment, United Nations Environment
  Programme, GRID-Arendal}, 2012.

\bibitem{Wiedenhoeft2019}
Wiedenhoeft~A. C., Simeone J, Smith A., Parker-Forney M, Soares R, and Fishman
  A.
\newblock Fraud and misrepresentation in retail forest products exceeds {U}.
  {S}. forensic wood science capacity.
\newblock {\em PLoS ONE}, 14(7), July 2019.

\bibitem{LoweSasaki2016}
Andrew~J. Lowe, Eleanor~E. Dormontt, Matthew~J. Bowie, Bernd Degen, Shelley
  Gardner, Darren Thomas, Caitlin Clarke, Anto Rimbawanto, Alex Wiedenhoeft,
  Yafang Yin, and Nophea Sasaki.
\newblock Opportunities for improved transparency in the timber trade through
  scientific verification.
\newblock {\em BioScience}, 66(11):990--998, 2016.

\bibitem{Dormontt2015}
Eleanor~E. Dormontt, Markus Boner, Birgit Braun, Gerhard Breulmann, Bernd
  Degen, Edgard Espinoza, Shelley Gardner, Phil Guillery, John~C. Hermanson,
  Gerald Koch, Soon~Leong Lee, Milton Kanashiro, Anto Rimbawanto, Darren
  Thomas, Alex~C. Wiedenhoeft, Yafang Yin, Johannes Zahnen, and Andrew~J. Lowe.
\newblock Forensic timber identification: It's time to integrate disciplines to
  combat illegal logging.
\newblock {\em Biological Conservation}, 191:790--798, 2015.

\bibitem{HermansonWiedenhoeft2011}
J.~C. Hermanson and A.~C. Wiedenhoeft.
\newblock A brief review of machine vision in the context of automated wood
  identification systems.
\newblock {\em IAWA Journal}, 32(2):233--250, 2011.

\bibitem{TexClassSurvey2018}
Li~Liu, Jie Chen, Paul~W. Fieguth, Guoying Zhao, Rama Chellappa, and Matti
  Pietik{\"{a}}inen.
\newblock A survey of recent advances in texture representation.
\newblock {\em CoRR}, abs/1801.10324, 2018.

\bibitem{CimpoiDeepTexture2015}
Mircea Cimpoi, Subhransu Maji, and Andrea Vedaldi.
\newblock Deep filter banks for texture recognition and segmentation.
\newblock In {\em Proceedings of the IEEE Conference on Computer Vision and
  Pattern Recognition}, pages 3828--3836, 2015.

\bibitem{EstebanANN2009}
Luis~García Esteban, Francisco~García Fernández, Paloma de~Palacios~de
  Palacios, Ruth~Moreno Romero, and Nieves~Navarro Cano.
\newblock Artificial neural networks in wood identification: The case of two
  {J}uniperus species from the {C}anary {I}slands.
\newblock {\em IAWA Journal}, 30(1):87 -- 94, 2009.

\bibitem{Khalid2008}
Marzuki Khalid, E~Lew~Yi Lee, Rubiyah Yusof, and Miniappan Nadaraj.
\newblock Design of an intelligent wood species recognition system.
\newblock {\em International Journal of Simulation System, Science and
  Technology}, 9(3):9--19, 2008.

\bibitem{Haralick1979}
R.~M. {Haralick}.
\newblock Statistical and structural approaches to texture.
\newblock {\em Proceedings of the IEEE}, 67(5):786--804, May 1979.

\bibitem{AfricaMicroID2017}
Núbia Rosa, Maaike De~Ridder, Jan Baetens, Jan Van~den Bulcke, Mélissa
  Rousseau, Odemir Bruno, Hans Beeckman, Joris Van~Acker, and Bernard De~Baets.
\newblock Automated classification of wood transverse cross-section
  micro-imagery from 77 commercial {C}entral-{A}frican timber species.
\newblock {\em Annals of Forest Science}, 74:30, 06 2017.

\bibitem{LeCun1989}
Y.~LeCun, B.~Boser, J.~S. Denker, D.~Henderson, R.~E. Howard, W.~Hubbard, and
  L.~D. Jackel.
\newblock Backpropagation applied to handwritten zip code recognition.
\newblock {\em Neural Computation}, 1(4):541--551, Dec 1989.

\bibitem{Filho2014}
Pedro L.~Paula Filho, Luiz~S. Oliveira, Silvana Nisgoski, and Alceu~S. Britto.
\newblock Forest species recognition using macroscopic images.
\newblock {\em Machine Vision and Applications}, 25(4):1019--1031, May 2014.

\bibitem{CostaRicaCutting2018}
G.~{Figueroa-Mata}, E.~{Mata-Montero}, J.~C. {Valverde-Otárola}, and
  D.~{Arias-Aguilar}.
\newblock Evaluating the significance of cutting planes of wood samples when
  training {CNN}s for forest species identification.
\newblock In {\em 2018 IEEE 38th Central America and Panama Convention
  (CONCAPAN XXXVIII)}, pages 1--5, Nov 2018.

\bibitem{Ravindran2018}
Prabu Ravindran, Adriana Costa, Richard Soares, and Alex~C. Wiedenhoeft.
\newblock Classification of {CITES}-listed and other neotropical {M}eliaceae
  wood images using convolutional neural networks.
\newblock {\em Plant Methods}, 14(1), Mar 2018.

\bibitem{PanTransferLearnSurvey2010}
S.~J. Pan and Q.~Yang.
\newblock A survey on transfer learning.
\newblock {\em IEEE Transactions on Knowledge and Data Engineering},
  22(10):1345--1359, Oct 2010.

\bibitem{ZamirTaskonomy2018}
Amir~R Zamir, Alexander Sax, William Shen, Leonidas~J Guibas, Jitendra Malik,
  and Silvio Savarese.
\newblock Taskonomy: Disentangling task transfer learning.
\newblock In {\em Proceedings of the IEEE Conference on Computer Vision and
  Pattern Recognition}, pages 3712--3722, 2018.

\bibitem{Simonyan2014}
Karen Simonyan and Andrew Zisserman.
\newblock Very deep convolutional networks for large-scale image recognition.
\newblock {\em CoRR}, abs/1409.1556, 2014.

\bibitem{HeResNet2015}
Kaiming He, Xiangyu Zhang, Shaoqing Ren, and Jian Sun.
\newblock Deep residual learning for image recognition.
\newblock {\em CoRR}, abs/1512.03385, 2015.

\bibitem{XyloScope}
Hermanson J.C., Dostal D., Destree J.C., and A.C. Wiedenhoeft.
\newblock The {X}ylo{S}cope – a field deployable macroscopic digital imaging
  device for wood.
\newblock Technical Report FPL-RN-0367, U.S. Department of Agriculture, Forest
  Service, Forest Products Laboratory, 2019.

\bibitem{Russakovsky2015}
Olga Russakovsky, Jia Deng, Hao Su, Jonathan Krause, Sanjeev Satheesh, Sean Ma,
  Zhiheng Huang, Andrej Karpathy, Aditya Khosla, Michael Bernstein,
  Alexander~C. Berg, and Li~Fei-Fei.
\newblock {I}mage{N}et large scale visual recognition challenge.
\newblock {\em International Journal of Computer Vision}, 115(3):211--252, Dec
  2015.

\bibitem{IoffeSzegedyBN2015}
Sergey Ioffe and Christian Szegedy.
\newblock Batch normalization: Accelerating deep network training by reducing
  internal covariate shift.
\newblock {\em CoRR}, abs/1502.03167, 2015.

\bibitem{HintonDropout2012}
Geoffrey~E. Hinton, Nitish Srivastava, Alex Krizhevsky, Ilya Sutskever, and
  Ruslan Salakhutdinov.
\newblock Improving neural networks by preventing co-adaptation of feature
  detectors.
\newblock {\em CoRR}, abs/1207.0580, 2012.

\bibitem{Nair2010}
Vinod Nair and Geoffrey~E. Hinton.
\newblock Rectified linear units improve {R}estricted {B}oltzmann {M}achines.
\newblock In {\em Proceedings of the 27th International Conference on
  International Conference on Machine Learning}, pages 807--814, 2010.

\bibitem{HeInit2015}
Kaiming He, Xiangyu Zhang, Shaoqing Ren, and Jian Sun.
\newblock Delving deep into rectifiers: Surpassing human-level performance on
  {I}mage{N}et classification.
\newblock In {\em Proceedings of the IEEE international conference on computer
  vision}, pages 1026--1034, 2015.

\bibitem{Razavian2014}
Ali~Sharif Razavian, Hossein Azizpour, Josephine Sullivan, and Stefan Carlsson.
\newblock {CNN} features off-the-shelf: an astounding baseline for recognition.
\newblock {\em CoRR}, abs/1403.6382, 2014.

\bibitem{KingmaAdamOpt2014}
Diederik~P. Kingma and Jimmy Ba.
\newblock Adam: {A} method for stochastic optimization.
\newblock {\em CoRR}, abs/1412.6980, 2014.

\bibitem{SmithDisciplined2018}
Leslie~N. Smith.
\newblock A disciplined approach to neural network hyper-parameters: Part 1 -
  learning rate, batch size, momentum, and weight decay.
\newblock {\em CoRR}, abs/1803.09820, 2018.

\bibitem{howard2018fastai}
Jeremy Howard et~al.
\newblock fastai.
\newblock \url{https://github.com/fastai/fastai}, 2018.

\bibitem{SmithCyclical2017}
L.~N. Smith.
\newblock Cyclical learning rates for training neural networks.
\newblock In {\em 2017 IEEE Winter Conference on Applications of Computer
  Vision (WACV)}, pages 464--472, Los Alamitos, CA, USA, Mar 2017. IEEE
  Computer Society.

\bibitem{Cutout2017}
Terrance Devries and Graham~W. Taylor.
\newblock Improved regularization of convolutional neural networks with cutout.
\newblock {\em CoRR}, abs/1708.04552, 2017.

\bibitem{Scipy2014}
Eric Jones, Travis Oliphant, and Pearu Peterson.
\newblock {SciPy}: Open source scientific tools for {Python}.
\newblock {\em www.scipy.org}, 2014.

\bibitem{Recht2018}
Benjamin Recht, Rebecca Roelofs, Ludwig Schmidt, and Vaishaal Shankar.
\newblock Do {CIFAR-10} classifiers generalize to {CIFAR-10}?
\newblock {\em CoRR}, abs/1806.00451, 2018.

\bibitem{Recht2019}
Benjamin Recht, Rebecca Roelofs, Ludwig Schmidt, and Vaishaal Shankar.
\newblock Do imagenet classifiers generalize to imagenet?
\newblock {\em CoRR}, abs/1902.10811, 2019.

\bibitem{CostaRicaXyl2018}
G.~{Figueroa-Mata}, E.~{Mata-Montero}, J.~C. {Valverde-Otárola}, and
  D.~{Arias-Aguilar}.
\newblock Automated image-based identification of forest species: Challenges
  and opportunities for 21st century xylotheques.
\newblock In {\em 2018 IEEE International Work Conference on Bioinspired
  Intelligence (IWOBI)}, pages 1--8, July 2018.

\bibitem{KochOneShot2015}
Gregory Koch, Richard Zemel, and Ruslan Salakhutdinov.
\newblock Siamese neural networks for one-shot image recognition.
\newblock In {\em ICML Deep Learning Workshop}, 2015.

\bibitem{VinyalsOneShot2016}
Oriol Vinyals, Charles Blundell, Timothy Lillicrap, koray kavukcuoglu, and Daan
  Wierstra.
\newblock Matching networks for one shot learning.
\newblock In D.~D. Lee, M.~Sugiyama, U.~V. Luxburg, I.~Guyon, and R.~Garnett,
  editors, {\em Advances in Neural Information Processing Systems 29}, pages
  3630--3638. Curran Associates, Inc., 2016.

\bibitem{ZemelFewShot2017}
Jake Snell, Kevin Swersky, and Richard Zemel.
\newblock Prototypical networks for few-shot learning.
\newblock In I.~Guyon, U.~V. Luxburg, S.~Bengio, H.~Wallach, R.~Fergus,
  S.~Vishwanathan, and R.~Garnett, editors, {\em Advances in Neural Information
  Processing Systems 30}, pages 4077--4087. Curran Associates, Inc., 2017.

\bibitem{Tay2017}
Xin~Jie Tang, Yong~Haur Tay, Nordahlia~Abdullah Siam, and Seng~Choon Lim.
\newblock Rapid and robust automated macroscopic wood identification system
  using smartphone with macro-lens.
\newblock {\em CoRR}, abs/1709.08154, 2017.

\end{thebibliography}
}

\end{document}